\documentclass[runningheads]{myllncs}
\usepackage[T1]{fontenc}
\usepackage{graphicx}
\usepackage{graphicx}
\usepackage{amsmath}
\usepackage{amssymb}
\usepackage{bm}
\usepackage{orcidlink}

\begin{document}
\title{DentalSplat: Dental Occlusion Novel View Synthesis from Sparse Intra-Oral Photographs}
\titlerunning{DentalSplat}

\author{
Yiyi Miao\inst{1,3}$^{*}$\orcidlink{0009-0008-4488-1272} \and
Taoyu Wu\inst{2,4}$^{\dagger,*}$\orcidlink{0009-0008-7991-6869} \and
Tong Chen\inst{1,3}\orcidlink{0000-0002-9107-6769} \and 
Sihao Li\inst{2,3}\orcidlink{0000-0003-4231-7455} \and
Ji Jiang\inst{6}\orcidlink{0009-0005-0935-5144} \and
Youpeng Yang\inst{5}\orcidlink{0000-0003-2248-8679} \and
Angelos Stefanidis\inst{1}\orcidlink{0000-0002-4703-8765} \and
Limin Yu\inst{2}\orcidlink{0000-0002-6891-0604}$^{\mathparagraph}$ \and
Jionglong Su\inst{1}\orcidlink{0000-0001-5360-6493}$^{\mathparagraph}$
}

\authorrunning{Y. Miao et al.}
%
\institute{School of AI and Advanced Computing, Xi'an Jiaotong Liverpool University, Suzhou, China\\
\email{\{Yiyi.Miao21, Tong.Chen19\}@student.xjtlu.edu.cn, \{Jionglong.Su, Angelos Stefanidis\}@xjtlu.edu.cn}
\and
School of Advanced Technology, Xi'an Jiaotong Liverpool University, Suzhou, China\\
\email{\{Taoyu.Wu21, Sihao.li19\}@student.xjtlu.edu.cn, Limin.yu@xjtlu.edu.cn}
\and
School of Electrical Engineering, Electronics and Computer Science
, University of Liverpool, Liverpool, United Kingdom\\
\and
School of Physical Sciences, University of Liverpool, Liverpool, United Kingdom\\
\and
College of Computer Science and Technology, Zhejiang University, Hangzhou, China\\
\email{ypyang@zju.edu.cn}
}
\maketitle

\renewcommand\thefootnote{}

\footnotetext{${}^*$ Co-first authors.}
\footnotetext{${}^\dagger$ Project Lead.}
\footnotetext{${}^\mathparagraph$ Corresponding authors.}

\renewcommand\thefootnote{\arabic{footnote}}

\begin{abstract}
  In orthodontic treatment, particularly within telemedicine contexts, observing
  patients' dental occlusion from multiple viewpoints facilitates timely clinical
  decision-making. Recent advances in 3D Gaussian Splatting (3DGS) have shown
  strong potential in 3D reconstruction and novel view synthesis. However,
  conventional 3DGS pipelines typically rely on densely captured multi-view
  inputs and precisely initialized camera poses, limiting their practicality. Orthodontic cases, in contrast, often comprise only three sparse images, specifically, the anterior view and bilateral buccal views, rendering the reconstruction task especially challenging. The extreme sparsity of input views severely degrades
  reconstruction quality, while the absence of camera pose information further
  complicates the process. To overcome these limitations, we propose
  \textbf{DentalSplat}, an effective framework for 3D reconstruction from sparse
  orthodontic imagery. Our method leverages a prior-guided dense stereo
  reconstruction model to initialize the point cloud, followed by a
  scale-adaptive pruning strategy to improve the training efficiency and
  reconstruction quality of 3DGS. In scenarios with extremely sparse viewpoints,
  we further incorporate optical flow as a geometric constraint, coupled with
  gradient regularization, to enhance rendering fidelity. We validate our
  approach on a large-scale dataset comprising 950 clinical cases and an
  additional video-based test set of 195 cases designed to simulate real-world
  remote orthodontic imaging conditions. Experimental results demonstrate that
  our method effectively handles sparse input scenarios and achieves superior
  novel view synthesis quality for dental occlusion visualization, outperforming
  state-of-the-art techniques.
\end{abstract}

\section{Introduction}
Accurate dental occlusion reconstruction~\cite{zhang2015reconstruction} is
crucial for orthodontic treatment, impacting planning, adjustments, and
long-term function. Traditionally, Cone Beam Computed Tomography
(CBCT)~\cite{kapila2011current} and Intraoral Scanning
(IOS)~\cite{elgarba2024ai} have been key tools, providing high-resolution 3D
models of teeth and their relationships. CBCT visualizes hard tissues like
teeth and jawbones, aiding occlusal analysis, while IOS captures surface
details of dental arches to create precise digital
models~\cite{mangano2017intraoral}. These are often combined with occlusion
registration, where the patient bites into a set position to align the
arches~\cite{xu2023occluded}. However, these methods require specialized
equipment and expertise~\cite{johnson2019digital}, limiting their use to remote
monitoring. Thus, an alternative approach is needed to achieve accurate
occlusal reconstruction with minimal input, reducing the reliance on advanced
imaging.

Artificial Intelligence (AI) technologies, such as
DentalMonitoring~\cite{impellizzeri2020dental} and Invisalign Virtual Care
AI~\cite {thurzo2021artificial}, have become integral tools in dental clinics
for remote orthodontic treatment monitoring~\cite{sharp2025orthodontics}. While
these systems enable patients to capture dental images using
smartphone-connected devices, they primarily rely on single-view images, which
significantly limit comprehensive assessments of occlusal relationships and
spatial positioning~\cite{hansa2021artificial}. Novel View Synthesis (NVS)
offers a promising solution to this limitation by generating new viewpoints
from a set of input images, thereby enabling more accurate occlusion
evaluation. Recently, 3D Gaussian Splatting (3DGS)~\cite{kerbl20233dgs} has
demonstrated superior performance in terms of rendering efficiency and
photorealism through its use of explicit 3D Gaussian representations and
differentiable rasterization~\cite{ulsr}. However, despite its excellent
rendering quality, 3DGS~\cite{kerbl20233dgs} heavily depends on
Structure-from-Motion (SfM) methods such as COLMAP for camera pose estimation
and initialization. This dependency presents a significant challenge for
real-world applications like remote oral diagnostics, where image acquisition
is inherently sparse. Several approaches have been proposed to reduce 3DGS's
reliance on dense image inputs. MVSplat~\cite{mvsplat} introduces a
Transformer-based framework that incorporates pre-trained geometric priors and
epipolar constraints to infer depth information and guide reconstruction.
Similarly, Nope-NeRF~\cite{nopenerf} and CF-3DGS~\cite{cf3dgs} utilize
depth-based constraints to minimize dependence on COLMAP for pose estimation.
Nevertheless, these methods typically assume overlapping views and continuous
video inputs, making them less suitable for truly sparse, pose-free scenarios
in novel view synthesis and scene reconstruction tasks. DUSt3R~\cite{dust3r}
addresses these limitations by requiring only two sparse, unposed images to
generate point and confidence maps for end-to-end 3D reconstruction. By
leveraging pre-trained models, DUSt3R produces high-quality 3D models directly
from RGB images while simultaneously providing camera poses. This approach
effectively supports various tasks, including intrinsic recovery and pose
estimation, making it particularly suitable for applications with limited input
data.

Despite addressing the limitations of SfM dependency in sparse input scenarios,
DUSt3R faces significant challenges in orthodontic
applications~\cite{xie2024tooth}. In remote orthodontic practice, patients
often use various mobile devices with varying camera qualities and inherent
noise, resulting in sparse and uncalibrated images that compromise accurate
visualization~\cite{snider2023effectiveness}. The inherent characteristics of
orthodontic imaging, including specular reflections from tooth enamel, motion
blur during intraoral image capture, and variable lighting
conditions~\cite{kalpana2018digital}, further impact the performance of the
3DGS pipeline. When utilizing DUSt3R for 3DGS initialization, the dense point
clouds generated by DUSt3R may reduce computational efficiency and hinder
convergence. The excessive density of these point clouds can significantly
degrade 3DGS optimization. Moreover, in standard orthodontic imaging protocols,
there are substantial angular differences between frontal and bilateral buccal
views. Without proper geometric constraints, this can lead to suboptimal
reconstruction and rendering quality. These challenges necessitate specialized
adaptations to effectively integrate DUSt3R with 3DGS for orthodontic
applications.

To address the challenges of sparse input and unknown camera poses in
orthodontic scenarios, we present \textbf{DentalSplat}, a novel 3D dental
reconstruction framework based on 3DGS and DUSt3R, designed to address the
challenges of sparse input and unknown camera poses in orthodontic scenarios.
This is the first framework capable of achieving high-quality novel view
synthesis and 3D reconstruction from sparse, pose-free dental images within a
minute. Specifically, we introduce a Scale-Adaptive Pruning (SAP) strategy for
Gaussian Splatting, which operates on the dense point clouds generated by
DUSt3R. This strategy analyzes the spatial distribution characteristics of
point clouds to determine adaptive thresholds for different spatial regions,
effectively handling both dense and sparse areas. This approach significantly
reduces the initial 3D point cloud size while maintaining quality, thereby
decreasing both optimization time and computational overhead in 3DGS. To
address the challenges of reflections and motion blur in dental reconstruction,
we incorporate optical flow constraints by computing the residual between the
optical flow generated from 3DGS projections of adjacent frames and that from
the original 2D images. This optical flow loss is integrated with the
traditional photometric loss to enhance multi-view geometric consistency.
Furthermore, to mitigate the blurring artifacts that can occur in 3DGS due to
inaccurate gradients affecting splitting and cloning operations during
optimization, we compute gradient weights for each Gaussian, effectively
reducing local over-reconstruction artifacts.

Our main contributions can be summarized in threefold:
\begin{itemize}
  \item We enhance the SAP strategy to mitigate the computational burden imposed by
        DUSt3R's dense point clouds during 3DGS optimization.
  \item We propose an enhanced differential Gaussian rasterization module with optical
        flow and gradient-weighted optimization, effectively improving the rendering
        quality of complex dental structures.
  \item We validate our framework on a self-collected dataset of 956 clinical dental
        cases, demonstrating superior reconstruction speed and novel view synthesis
        quality under sparse input conditions compared to baseline methods.
\end{itemize}

\section{Related Work}

\subsection{3D Scene Reconstruction}
For decades, 3D reconstruction from images has been dominated by classical
pipelines combining SfM and Multi-View Stereo (MVS). SfM systems, such as the
widely-used COLMAP~\cite{colmap}, first recover a sparse 3D point cloud and
camera poses by matching local features across multiple views and performing
bundle adjustment. Subsequently, MVS algorithms densify this sparse
representation by leveraging photometric consistency across views. A paradigm
shift occurred with the introduction of Neural Radiance Fields
(NeRF)~\cite{mildenhall2021nerf}, which represents a scene as a continuous 5D
function learned by a Multi-Layer Perceptron (MLP). By mapping 3D coordinates
and a 2D viewing direction to volume density and color, NeRF achieves
state-of-the-art photorealism for novel view synthesis through differentiable
volume rendering. However, the original NeRF is slow to train and render, and
critically, it requires a dense set of input images with accurate camera poses,
typically pre-computed using COLMAP. Subsequent research has focused on
mitigating these limitations. Mip-NeRF~\cite{barron2021mip} addressed aliasing
artifacts by rendering anti-aliased conical frustums instead of rays, improving
detail representation across different scales. More recently,
3DGS~\cite{kerbl20233dgs} has emerged as a leading method, combining the
benefits of explicit representations with the differentiability of neural
rendering. 3DGS models a scene as a collection of 3D Gaussians, each with
optimizable properties such as position, covariance, color, and opacity.

\subsection{Camera Pose-Free Reconstruction}
A significant research thrust has focused on eliminating the reliance on
pre-computed camera poses from SfM. These methods aim to jointly optimize the
scene representation and camera parameters. BARF~\cite{lin2021barf} was a
pioneering work that enabled the joint optimization of camera poses and a NeRF
model. It introduced a coarse-to-fine registration strategy by gradually
unmasking high-frequency components of the positional encoding, which proved
crucial for avoiding poor local minima. Nope-NeRF~\cite{nopenerf} incorporates
geometric priors from a monocular depth estimator to constrain the relative
poses between frames, stabilizing the joint optimization process. The pose-free
paradigm has also been extended to 3D Gaussian Splatting. CF-3DGS~\cite{cf3dgs}
adapts 3DGS for video streams without SfM pre-processing by sequentially
estimating the relative pose of each new frame and progressively growing the
set of Gaussians. This approach leverages the temporal continuity of video
input and the explicit nature of the Gaussian representation to achieve robust
tracking and reconstruction.

\subsection{Sparse-View Reconstruction}
Another critical challenge is reconstructing scenes from a sparse set of input
views, where per-scene optimization is highly under-constrained and prone to
overfitting. PixelNeRF~\cite{yu2021pixelnerf} conditions a NeRF on image
features extracted by a convolutional network, allowing it to synthesize novel
views from a single image in a feed-forward pass.
MVSNeRF~\cite{chen2021mvsnerf} integrates principles from MVS by constructing a
plane-swept cost volume from as few as three views, providing a powerful
geometric prior that enables high-quality generalization. For sparse-view 3DGS,
MVSplat~\cite{chen2024mvsplat} leverages a plane-swept cost volume to infer
geometric cues from multi-view stereo, which then guides the direct,
feed-forward prediction of 3D Gaussian parameters. This geometry-aware approach
demonstrates strong generalization and efficiency for sparse inputs.

\subsection{End-to-End Reconstruction from Unposed Images}
The DUSt3R model~\cite{dust3r} addresses the challenges of sparse inputs and
unknown poses by enabling end-to-end 3D reconstruction from unposed,
uncalibrated image pairs. It predicts relative camera poses and dense depth
maps, acting as a general-purpose geometric foundation model. DUSt3R eliminates
the need for traditional Structure from Motion (SfM) pipelines by leveraging a
large, diverse training dataset. However, DUSt3R's limitation lies in its
pairwise input processing, which introduces computational inefficiencies,
especially with larger image sets. To address this,
MUSt3R~\cite{cabon2025must3r} extends DUSt3R to multi-view reconstruction,
allowing all views to be processed in a single forward pass.

Despite the power of these general-purpose models, as we identify in our work,
their direct application for initializing 3DGS in specialized domains like
orthodontics presents unique challenges, such as the computational burden of
dense point cloud outputs and susceptibility to domain-specific artifacts.
These limitations motivate our proposed contributions in DentalSplat, which adapt
and refine this powerful prior for high-fidelity dental reconstruction.

\section{Methodology}
\begin{figure}[htb]
  \centering
  \includegraphics[width=\textwidth, height=\textheight, keepaspectratio]{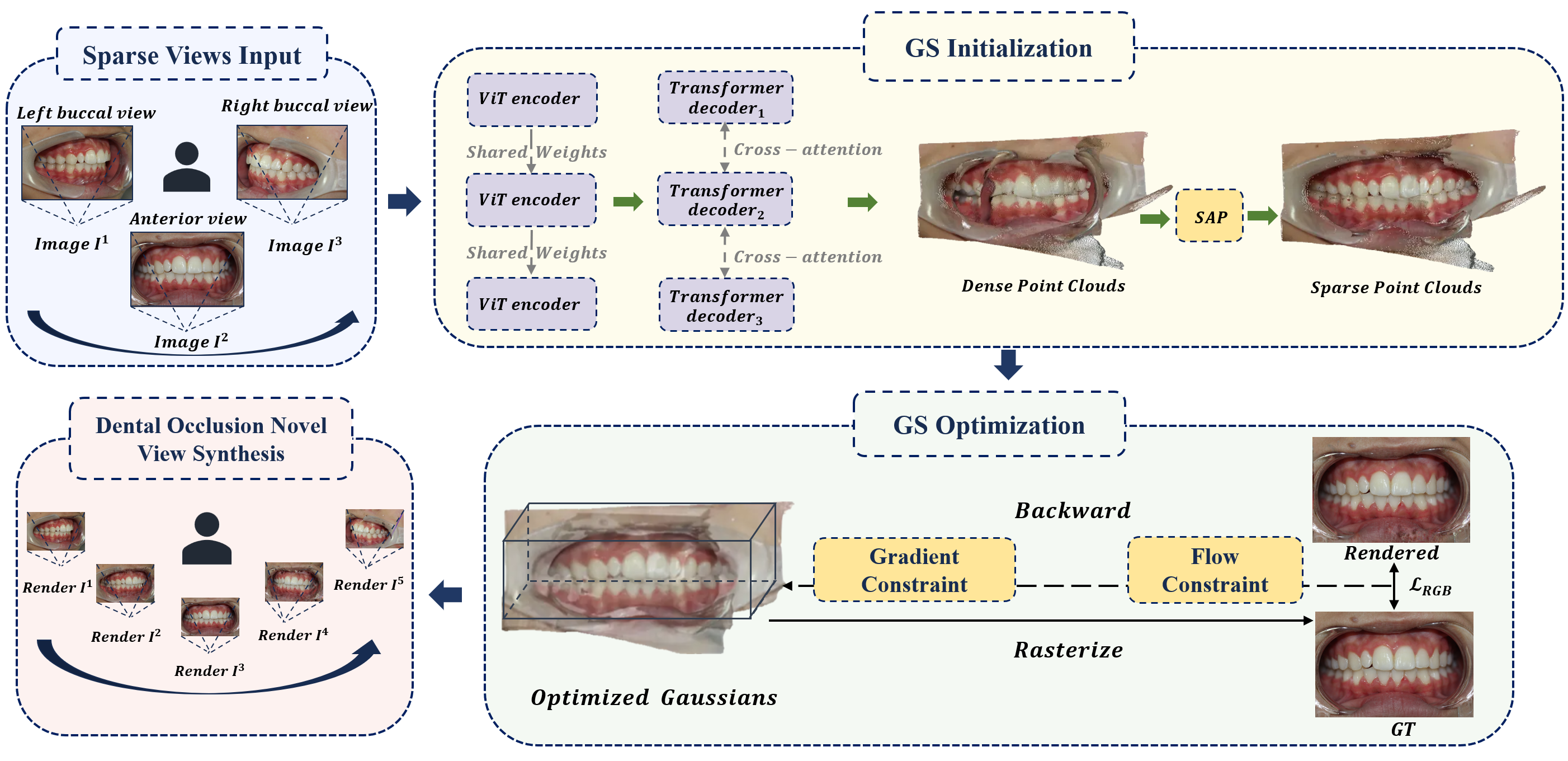}
  \caption{\textbf{Overview of DentalSplat.} Given a set of sparse and unposed input images,  we leverage a stereo-dense reconstruction model to regress the dense point cloud of these input images in the global coordinate system and obtain the corresponding relative camera pose. Subsequently, we apply the SAP strategy to eliminate outlier points, followed by downsampling to obtain a sparse point cloud suitable for 3DGS initialization. During optimization, we incorporate optical flow constraints to ensure geometric consistency and employ gradient constraints to enhance the densification of the 3DGS.}
  \label{fig: overview}
\end{figure}

\subsection{Preliminary}
\textbf{3D Gaussian Splatting.}
3DGS~\cite{kerbl20233dgs} represents a scene using an explicit collection of anisotropic Gaussian primitives defined in 3D space. Each primitive $G_i$ is parametrized by a mean position $\boldsymbol{\mu}_i \in \mathbb{R}^3$, an opacity $o_i \in [0,1]$, and a covariance matrix $\boldsymbol{\Sigma}_i \in \mathbb{R}^{3 \times 3}$. The spatial density of each Gaussian is given as:
\begin{equation}
  G_i(\mathbf{X}) = o_i \cdot \exp\left\{-\frac{1}{2} (\mathbf{X} - \boldsymbol{\mu}_i)^\top \boldsymbol{\Sigma}_i^{-1} (\mathbf{X} - \boldsymbol{\mu}_i)\right\},
\end{equation}
where $\mathbf{X} \in \mathbb{R}^3$ denotes an arbitrary 3D point. The covariance matrix $\boldsymbol{\Sigma} \in \mathbb{R}^{3 \times 3}$ can be decomposed into a scaling matrix and a rotation quaternion for efficient optimization.

The rendering process projects these 3D Gaussians onto the 2D image plane for a
given camera pose. The final color $\hat{C}(\mathbf{p})$ and depth
$\hat{D}(\mathbf{p})$ for each pixel $\mathbf{p}$ are synthesized by
alpha-blending the contributions of all Gaussians that overlap with the pixel,
sorted from front to back along the camera ray. The blending process is
formulated as:
\begin{equation}
  \hat{C}(\mathbf{p}) = \sum_{i \in \mathcal{N}} c_i \alpha_i \prod_{j=1}^{i-1}(1-\alpha_j), \quad \hat{D}(\mathbf{p}) = \sum_{i \in \mathcal{N}} d_i \alpha_i \prod_{j=1}^{i-1}(1-\alpha_j),
\end{equation}
where $\mathcal{N}$ is the set of sorted Gaussians. For each Gaussian $i$, $c_i$ is its color, determined by its SH coefficients for the current viewing direction, and $d_i$ is its depth, corresponding to the z-coordinate of its center $\boldsymbol{\mu}_i$ in camera space. The blending weight, $\alpha_i$, is crucial and is calculated by multiplying the learned opacity $o_i$ with the value of the projected 2D Gaussian's probability density function evaluated at the pixel center $\mathbf{p}$. This formulation allows for differentiable rendering, enabling end-to-end optimization of the Gaussian attributes through gradient-based methods.

\vspace{1pt}
\noindent\textbf{Dust3R.}
Dust3R~\cite{dust3r} proposes a unified and calibration-free 3D reconstruction framework that bypasses the traditional reliance on keypoint correspondences and explicit camera parameters. It introduces the concept of a \emph{pointmap}, a dense mapping from image pixels to 3D coordinates. Given an $i \times j$ RGB image $I$ and its corresponding depth map $D$, the pointmap $X \in \mathbb{R}^{W \times H \times 3}$ is computed in the camera coordinate system using the intrinsic matrix $K$ as follows:
\begin{equation}
  X_{i,j} = K^{-1} \begin{bmatrix} i D_{i,j}, j D_{i,j}, D_{i,j} \end{bmatrix}^T.
\end{equation}

Given two views $I_{t_1}$ and $I_{t_2}$, their respective pointmaps $X_{t_1},
  X_{t_2}$ can be aligned via a rigid transformation:
\begin{equation}
  X_{t_1 \to t_2} = T_{t_2} T_{t_1}^{-1} X_{t_1},
\end{equation}
where $T_{t_1}, T_{t_2} \in SE(3)$ are the world-to-camera transformations for each view.

The training objective of Dust3R is based on 3D regression with scale
normalization and confidence-aware optimization. For each valid pixel $ i $ in
frames $ I_{t1} $ and $ I_{t2} $, the regression loss is computed as the
Euclidean distance between the predicted and ground-truth point maps, scaled to
resolve scale ambiguity:
\begin{equation}
  \mathcal{L}_{\text{reg}}(v,i) = \left\| \frac{1}{z} X_i^{v} - \frac{1}{z'} X_i^{gt} \right\|.
\end{equation}

To address scale ambiguity, Dust3R normalizes the predicted and ground-truth
point maps using scaling factors $ z $ and $ z' $, which represent the average
distance of valid points from the origin:
\begin{equation}
  z = \frac{1}{|D_1| + |D_2|} \sum_{i \in D} \| X_i \|.
\end{equation}

Dust3R also introduces a confidence-aware loss to mitigate issues arising from
poorly defined 3D points, such as those in the sky or translucent objects. The
confidence score for each pixel $C^i_v$ is defined as $1 +
  \exp(\tilde{C}^i_v)$, ensuring positivity and enabling adaptive loss weighting:
\begin{equation}
  \mathcal{L}_{\text{conf}} = \sum_{v \in \{1,2\}} \sum_{i \in \mathcal{D}^v} C^i_v \left\| \frac{1}{z} \hat{X}^v_i - \frac{1}{z'} X^{\text{gt}}_i \right\| - \alpha \log C^i_v.
  \label{eq:conf_loss}
\end{equation}
Equation~\eqref{eq:conf_loss} promotes robustness to geometric ambiguities and provides a means to infer per-pixel confidence, which can be utilized in downstream tasks such as global alignment and visual localization. Dust3R's output pointmaps serve as a strong initialization for 3DGS and enable consistent 3D representations without requiring extrinsic or intrinsic camera calibration.

\subsection{3DGS Initialization}
\textbf{Initialization.}
For sparse and unposed orthodontic input images, we employ DUSt3R to generate a point cloud that serves as the initialization for 3DGS training. Specifically, once the DUSt3R network is optimized, it produces precise point maps for the given frames. These point maps enable the recovery of camera parameters and globally aligned point clouds, effectively resolving the convergence issues encountered by COLMAP with sparse, uncalibrated images. The point clouds derived from DUSt3R provide a robust foundation for initializing 3DGS primitives.

\vspace{1pt}
\noindent \textbf{Scale-Adaptive Pruning.}
Although DUSt3R generates dense point clouds, their excessive density can adversely impact 3DGS optimization efficiency and scene representation quality. Unlike the original 3DGS approach, which relies on sparse point clouds from COLMAP, we propose the SAP method, which necessitates additional pruning to make the dense DUSt3R point clouds compatible with the Adaptive Density Control (ADC) mechanism. Inspired by~\cite{instantsplat,absgs}, we implement an efficient pruning strategy that filters the initial point cloud after downsampling. This approach selectively retains the most significant points based on their spatial influence, as characterized by their scaling parameters.

We denote the scaling parameters of all Gaussians as $ \mathbf{S} \in
  \mathbb{R}^{N \times 3} $, where $ \mu_S $ represents the mean magnitude of all
scaling components. The pruning masks are computed as:
\begin{equation}
  \begin{aligned}
    \mathcal{M}_1 & = \left\{ \mathbf{S}_i \mid \max(\mathbf{S}_i) > \mu_S \right\} \\
    \mathcal{M}_2 & =
    \begin{cases}
      \left\{ \mathbf{S}_i \mid \max(\mathbf{S}_i) > Q_i(\mathbf{S}) \right\} & \text{if } N < 5 \times 10^6, \\
      \left\{ \mathbf{S}_i \mid \max(\mathbf{S}_i) > 4 \mu_S \right\}         & \text{otherwise},
    \end{cases}
  \end{aligned}
\end{equation}
where $Q_{i}$ is $(\cdot)$ denotes the percentile quantile function.
The final pruning mask is obtained through a logical conjunction: $
  \mathcal{M}_{\text{final}} = \mathcal{M}_1 \cap \mathcal{M}_2. $

The pruning method removes outliers while retaining critical points in complex
geometries. Adjusting the threshold based on point cloud size prevents
excessive pruning in large scenes and ensures robust outlier removal in smaller
ones. The surviving Gaussians meet the condition $
  \mathcal{G}_{\text{survived}} = \{\mathbf{G}_i \mid \mathbf{S}_i \in
  \mathcal{M}_{\text{final}}\}, $ where $ \mathbf{G}_i $ represents the $ i $-th
Gaussian primitive.

\subsection{3DGS Optimization}
\subsubsection{Gradient Constraint.}
The Gradient Collision issue represents a significant challenge in 3DGS,
manifesting as poor reconstruction quality and regional
blur~\cite{mallick2024taming,ulsr}. This issue causes conflicts between gradient
directions from different pixels. Each 3DGS influences multiple pixels, and
each pixel is affected by multiple 3DGS elements. When gradients conflict, the
accumulated gradient for a 3DGS weakens, hindering densification operations.

To simplify the notion, consider a single 3DGS element $G_i$ projected onto the
2D image plane as a 2D Gaussian $g_i$ centered at $\mu_i$, affecting $n$
pixels. The total loss function $L$ quantifies the discrepancy between
predicted and actual values, with gradients calculated as $\sum_{j=1}^n\frac{\partial L_j}{\partial \mu_{i,x}}$ and $\sum_{j=1}^n\frac{\partial L_j}{\partial \mu_{i,y}}$, where $n$ denotes the number of pixels affected by $g_i$, and
$L_j$ represents the loss computed for the $j$-th pixel. Significant variation
in pixel gradient directions leads to Gradient Collision, causing gradient
accumulation to decrease. This misalignment prevents accurate splitting
direction estimation, resulting in ineffective splits and increased blur in
over-reconstructed regions.

To address this issue, we conduct an absolute operation to constrain the
gradients following~\cite{absgs}. The absolute operation method aligns gradient
directions along their axes, ensuring consistency. By mitigating Gradient
Collision, it reduces blur, especially in over-reconstructed regions. The
absolute operation is defined as:
\begin{equation}
  \hat{g}_{i,x} = \sum_{j=1}^n \left| \frac{\partial L_j}{\partial \mu_{i,x}} \right|, \quad \hat{g}_{i,y} = \sum_{j=1}^n \left| \frac{\partial L_j}{\partial \mu_{i,y}} \right|.
\end{equation}

\vspace{1pt}
\noindent\textbf{Per-Gaussian Pixel Flow.}
To enhance rendering quality, we incorporate optical flow loss as a geometric constraint. Optical flow results from both object and camera motion, with camera movement being the primary contributor in our scenarios involving sparse input data and significant viewpoint differences~\cite{beauchemin1995computation,endoflow-slam}. In 3D Gaussians Splatting processing, each pixel $\mathbf{x}_{i}$ corresponds to a set of 3DGS, where the pixel colour is obtained by alpha-blending the 2D Gaussians projected from multiple 3D Gaussians. Building upon the work presented in ~\cite{gaussianflow}, at time $t$, we render the $i$-th 3D Gaussian using the camera pose $\mathcal{T}_{t}$ onto the 2D image plane, resulting in pixel $\mathbf{x}_{i,t}$. This pixel is mapped to the canonical space using the mean $\boldsymbol{\mu}_{i,t}$ and covariance matrix $\mathbf{\Sigma}_{i,t}$ of the corresponding $i$-th 2D Gaussian. At time $t+1$, the the pixel position $\mathbf{x}_{i,t+1}$ is determined by projecting the 3DGS through the unknown-but-sought camera pose $\hat{\mathcal{T}}_{t+1}$, as expressed by:
\begin{equation}
  \mathbf{x}_{i,t+1} = \pi\left(\mathcal{G}_{t},\mathcal{T}_{t+1}\right),
\end{equation}
where $\pi(\cdot)$ denotes the camera projection. From this, we can obtain the corresponding mean $\boldsymbol{\mu}_{i,t+1}$ and covariance matrix $\mathbf{\Sigma}_{i,t+1}$ for the $i$-th Gaussian. The Gaussian flow for the $i$-th Gaussian is given by the positional displacement, which represents the difference between the position of the pixel:
\begin{equation}
  \text{flow}_{i}^{G}(\mathbf{x}_{t}) = \mathbf{x}_{i,t+1} -\mathbf{x}_{i,t}.
\end{equation}
\begin{figure}[t]
  \centering
  \includegraphics[width=\textwidth, height=\textheight, keepaspectratio]{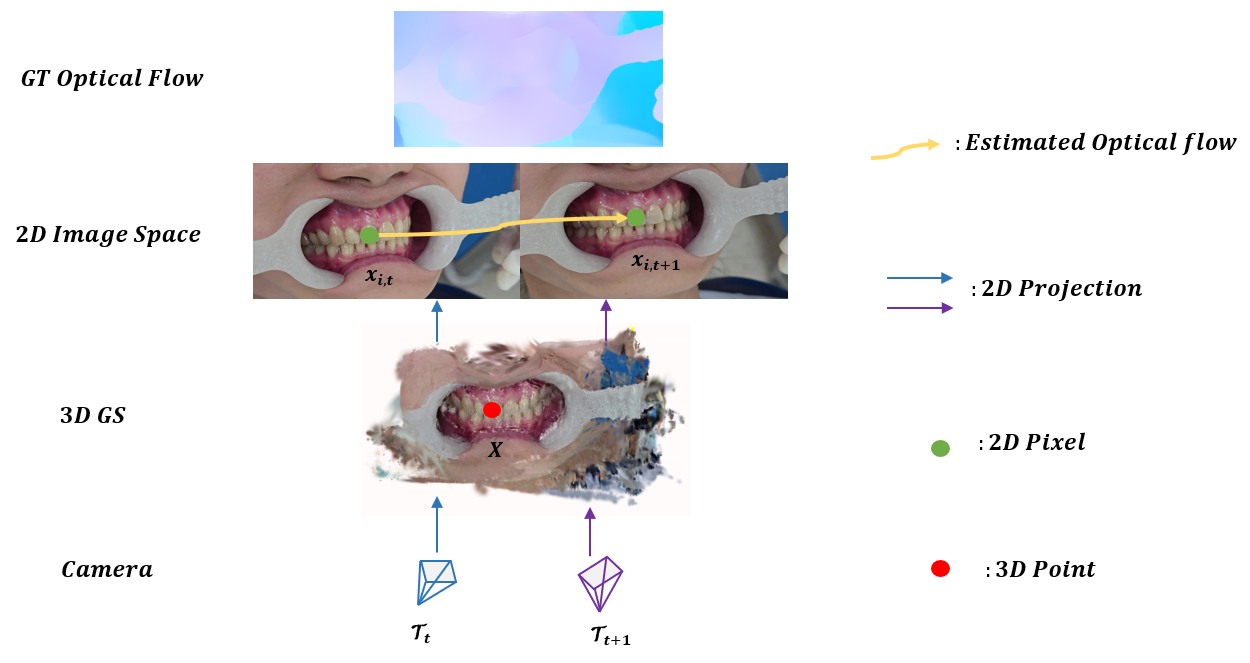}
  \caption{\textbf{Overview of Flow Constraint.} At time $t$, each 2D pixel $ x_t $ is formed by projecting $K$ overlapping 3D Gaussians under camera pose $\mathcal{T}_t$. At time $t+1$, their motions induce Gaussian flows whose projections are aggregated to estimate the overall optical flow. To jointly optimize the 3D Gaussian primitives $\hat{\mathcal{G}}$ and the camera pose $\mathcal{T}_{t+1}$, we minimize the residual between the estimated optical flow and the ground truth optical flow computed using an off-the-shelf method.}
  \label{fig: overview_flow}
\end{figure}

\vspace{1pt}
\noindent\textbf{Simultaneous Optimization by Flow constraint.} Unlike~\cite{gaussianflow}, in our work, the Gaussians are isotropic, where both covariance matrices are symmetric and positive definite. We jointly optimize the estimated camera pose $\hat{\mathcal{T}}_{t+1}$ and 3DGS primitive $\hat{\mathcal{G}}$ by the flow loss. Consequently, the Cholesky factorization ~\cite{cholesky} of the covariance matrices $\mathbf{\Sigma}_{i,t}$ and $\mathbf{\Sigma}_{i,t+1}$ simplifies to the identity matrix. This enables us to express the Gaussian flow for the $i$-th Gaussian equivalent to:
\begin{equation}
  \text{flow}_{i}^{G}(\mathbf{x}_{t}) = \boldsymbol{\mu}_{i,t+1} - \boldsymbol{\mu}_{i,t}.
\end{equation}

For each pixel with $K$ overlapping Gaussians, we compute the composite flow
through alpha-weighted blending:
\begin{equation}
  \text{flow}^{G}(\mathcal{T}_{t+1}, \mathcal{G}_{t}) = \sum_{i=1}^{K} w_i (\boldsymbol{\mu}_{i,t+1} - \boldsymbol{\mu}_{i,t}),
\end{equation}
where $w_i$ denotes the normalized blending weight of the $i$-th Gaussian along the camera ray. For adjacent frames $\mathbf{I}_t$ and $\mathbf{I}_{t+1}$, we obtain the optical flow $\text{flow}^{\mathrm{Gt}}(\mathbf{x})$ using an off-the-shelf method as ground truth. We then define the flow loss aggregated over all pixels as:
\begin{equation}
  \mathcal{L}_{\text{flow}} = \|\text{flow}^{G}(\mathcal{T}_{t+1}, \mathcal{G}_{t}) - \text{flow}^{Gt}(\mathbf{x})\|_2.
\end{equation}

However, in sparse scenes, optical flow prediction tends to introduce more
noise. To address this, we employ a bidirectional optical flow model, which
computes the forward optical flow between Frame $t$ and Frame $t+1$, as well as
the backward optical flow from Frame $t+1$ to Frame $t$. By leveraging the
bidirectional optical flow process, we can obtain the corresponding optical
flow confidence mask, denoted as $M(\mathbf{x}_{t_1})$. The confidence mask is
then applied to both the forward optical flow and the ground truth flow to
compute the adjusted flow loss. The per-pixel flow loss is then calculated as:
\begin{equation}
  \mathcal{L}_{\text{flow}}(\mathbf{x}_{t_1}) = \|M(\mathbf{x}_{t_1}) \odot \text{flow}^{\mathrm{G}}(\mathbf{x}_{t_1}) - M(\mathbf{x}_{t_1}) \odot \text{flow}^{Gt}(\mathbf{x}_{t_1})\|_2,
\end{equation}
where $M(\mathbf{x}_{t_1})$ represents the confidence mask applied to the pixel $\mathbf{x}_{t_1}$, and $\odot$ denotes element-wise multiplication.

In the whole training process, we simultaneously optimize the estimated camera
pose $\hat{\mathcal{T}}_{t+1}$ and 3DGS primitive $\hat{\mathcal{G}}$ by
minimizing the following objective function:
\begin{equation}
  \hat{\mathcal{T}}_{t+1}, \hat{\mathcal{G}} = \underset{\mathcal{T}_{t+1}, \mathcal{G}}{\text{argmin}}\, \left( \lambda_1 \mathcal{L}_{\text{rgb}} + \lambda_2 \mathcal{L}_{\text{flow}} \right),
  \label{eq:keyframeop}
\end{equation}
where the RGB loss $\mathcal{L}_{rgb}$ measures the $\mathcal{L}_1$ residual between the rendered RGB image $\hat{I}_{t+1}$ (using pose $T_t$) and the ground truth image $I_{t+1}$:

\section{Experiments}
\subsection{Implementation Details}
\vspace{1pt}
\noindent\textbf{Dataset Description.}
To evaluate the accuracy and robustness of our framework, we conducted
comprehensive experiments on a clinical intra-oral dataset collected in
collaboration with professional dental hospitals. All images were captured by
certified orthodontists using a Canon EOS 700D camera equipped with a 100mm
macro lens and operated in forced flash mode to ensure consistent illumination
and minimize lighting variability.

The dataset comprises two distinct subsets designed for different experiments.
The first video dataset consists of 195 clinical cases, each recorded as
intra-oral video by professional orthodontists to simulate remote orthodontic
scenarios. Each video captures a continuous transition from the right buccal
view, through the frontal occlusal view, to the left buccal view. From each
video, we uniformly sampled 24 frames based on the frame rate and video
duration. These frames were then evenly divided into a training view set and a
test view set, each containing 12 images. During training, only the camera
poses and corresponding 2D images from the training set were provided as input.
For training views, we examined four different sparse view scenarios using 3,
6, 9, and 12 viewpoints, respectively, to analyze the framework's performance
under different input conditions. Once training was completed, the optimized 3D
model was used to render novel 2D views at the camera poses in the test set,
thereby assessing the quality of novel view synthesis. The second image dataset
contains 950 clinical cases, each consisting of only three intra-oral
photographs: one anterior occlusal view capturing the full dentition from the
front, and bilateral buccal views from the left and right sides. These cases
were selected from routine orthodontic records and serve to evaluate the
framework's capacity to reconstruct and synthesize novel views under extremely
sparse input conditions.

\vspace{1pt}
\noindent\textbf{Experimental Setup.}
We conduct all experiments and evaluations on a desktop computer equipped with
an Intel Core i9-13900KF CPU and an NVIDIA GeForce RTX 4090 GPU. We apply the
same set of hyperparameters to all cases in the dataset. For the 3D Gaussians,
we follow the default training parameters from the original Gaussian Splatting
implementation~\cite{kerbl20233dgs}. We use the Adam optimizer~\cite{adam} to
update the Gaussian parameters. To balance rendering efficiency and quality, we
set the number of training iterations to 2000.

\subsection{Evaluation results}
\textbf{Comparative Experiments on Video Test Dataset.}
To evaluate the novel view synthesis capabilities of our framework, we conducted comprehensive qualitative and quantitative comparisons on our video dataset against the original 3DGS framework and several state-of-the-art baselines, including 3DGS, CF-3DGS, and InstantSplat. As shown in Table~\ref{table:6_9view}, we report the average values of three metrics across 195 cases, Peak Signal-to-Noise Ratio (PSNR)~\cite{psnr}, Structural Similarity Index Measure (SSIM)~\cite{SSIM}, and Learned Perceptual Image Patch Similarity (LPIPS)~\cite{Lpips}. Our method achieves the best performance across all three metrics. Standard 3DGS fails to converge during optimization when initialized with sparse multi-view inputs, as indicated by the "-" entries in the table. It is only trainable under the 12-view setting, yet still exhibits substantially lower rendering quality compared to other methods. This highlights a fundamental limitation of conventional approaches when applied to dental imaging scenarios, where observations are often restricted and highly sparse.
\begin{table}[htb]
  \centering
  \caption{\textbf{Quantitative evaluation} on video test dataset.}
  \begin{tabular}{c c c c | c c c}
    \hline
    Algorithm                        & \multicolumn{3}{c|}{3 Training views} & \multicolumn{3}{c}{6 Training views}                                                                           \\
    \cline{2-7}
                                     & PSNR$\uparrow$                        & SSIM$\uparrow$                       & LPIPS$\downarrow$ & PSNR$\uparrow$ & SSIM$\uparrow$ & LPIPS$\downarrow$ \\
    \hline
    3DGS~\cite{kerbl20233dgs}        & -                                     & -                                    & -                 & -              & -              & -                 \\
    InstantSplat~\cite{instantsplat} & 23.81                                 & \textbf{0.826}                       & 0.304             & 27.01          & 0.863          & 0.268             \\
    CF-3DGS~\cite{cf3dgs}            & 15.32                                 & 0.748                                & 0.443             & 18.01          & 0.795          & 0.277             \\
    \textbf{Ours}                    & \textbf{23.96}                        & 0.822                                & \textbf{0.301}    & \textbf{28.41} & \textbf{0.872} & \textbf{0.247}    \\
  \end{tabular}
  \begin{tabular}{c c c c | c c c}
    \hline
    Algorithm                        & \multicolumn{3}{c|}{9 Training views} & \multicolumn{3}{c}{12 Training views}                                                                            \\
    \cline{2-7}
                                     & PSNR$\uparrow$                        & SSIM$\uparrow$                        & LPIPS$\downarrow$ & PSNR$\uparrow$  & SSIM$\uparrow$ & LPIPS$\downarrow$ \\
    \hline
    3DGS~\cite{kerbl20233dgs}        & -                                     & -                                     & -                 & 11.51           & 0.53           & 0.574             \\
    InstantSplat~\cite{instantsplat} & 28.656                                & 0.890                                 & 0.241             & 29.315          & 0.898          & 0.235             \\
    CF-3DGS~\cite{cf3dgs}            & 21.29                                 & 0.812                                 & 0.374             & 23.02           & 0.853          & 0.337             \\
    \textbf{Ours}                    & \textbf{29.363}                       & \textbf{0.891}                        & \textbf{0.237}    & \textbf{30.174} & 0.897          & \textbf{0.213}    \\
    \hline
  \end{tabular}
  \label{table:6_9view}
\end{table}

\begin{figure}[htb]
  \centering
  \includegraphics[width=\textwidth, keepaspectratio]{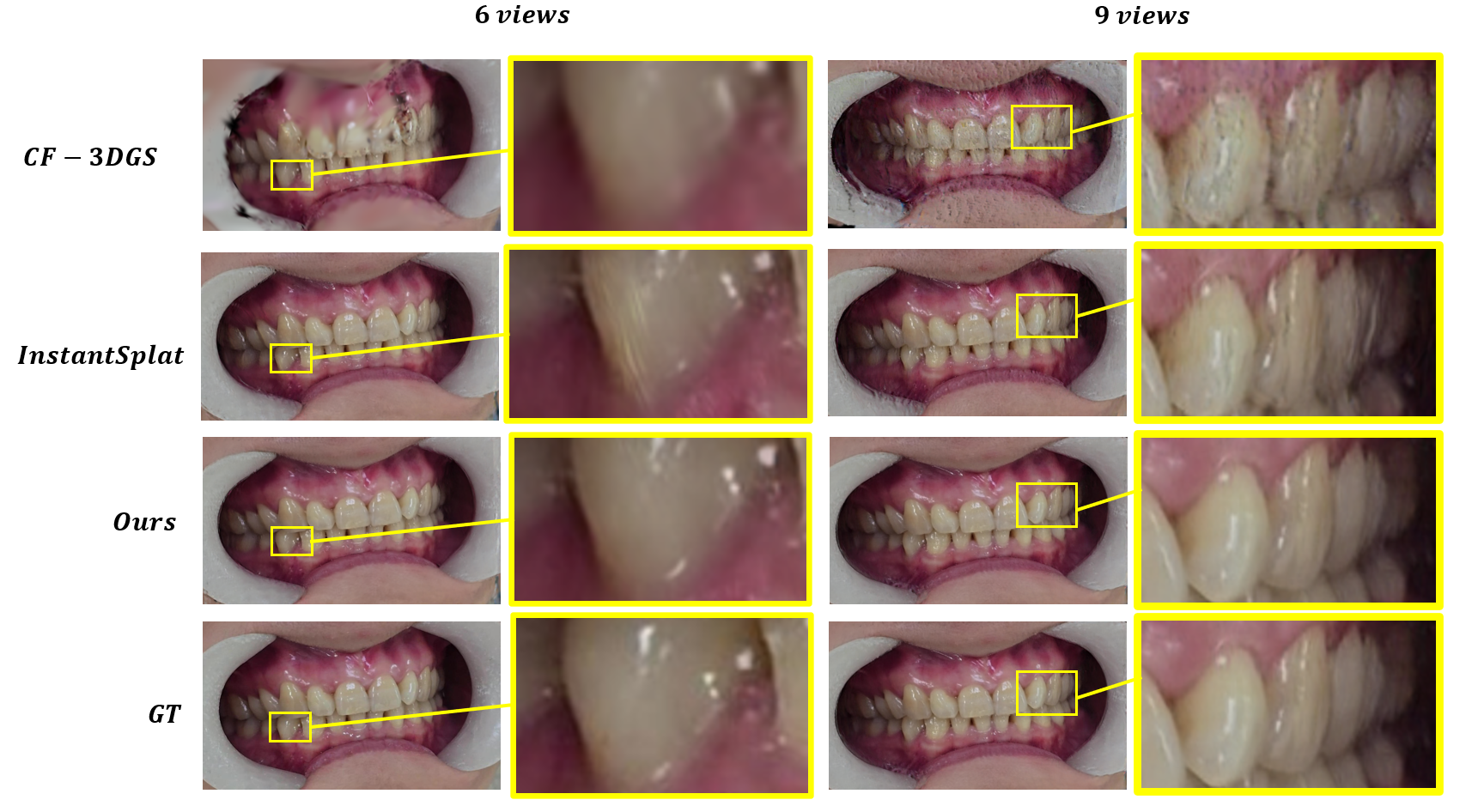}
  \caption{{\textbf{Novel View Synthesis Comparisons with 6 views and 9 views input.}We qualitatively compare the quality of novel view synthesis and show that our method has better quality with more accurate texture details.}}
  \label{fig: 69viewNovel}
\end{figure}

Figure~\ref{fig: 69viewNovel} illustrates the qualitative evaluations of novel
view synthesis. Under the 6-view input condition, CF-3DGS suffers from
noticeable blurring and floating artifacts. InstantSplat exhibits geometric
distortions in the lower teeth when compared with the ground truth. With 9-view
inputs, CF-3DGS eliminates major artifacts in the dental region but still
produces blurry and low-resolution images, with evident overfitting and
hallucinated geometry in the right buccal area. InstantSplat also suffers from
over-reconstruction in the right molars, leading to texture degradation and
shape distortion. These artifacts may adversely affect clinical assessment,
particularly in remote orthodontic follow-ups. In contrast, our reconstructions
remain artifact-free and preserve geometric fidelity across both 6-view and
9-view inputs, demonstrating strong generalization and high-quality novel view
synthesis performance.

\vspace{1pt}
\noindent \textbf{Comparative Experiments on 3 Views images Dataset.}
To further assess the framework's robustness under extremely sparse input conditions, we conducted additional experiments on an image dataset comprising 950 clinical cases. For each case, occlusal reconstruction was performed using only three intra-oral images: anterior, left buccal, and right buccal views. Since no test views are available in this dataset, Table~\ref{tab:3view} reports the reconstruction performance on the training views using the same three evaluation metrics (PSNR, SSIM, LPIPS) averaged over 956 cases. The standard 3DGS fails to converge under such sparse conditions, as denoted by "-".
\begin{table}[htb]
  \centering
  \caption{\textbf{Quantitative results} with the other methods using 3 views.}
  \begin{tabular}{c c c c c}
    \hline
    Methods                          & PSNR$\uparrow$ & SSIM$\uparrow$ & LPIPS$\downarrow$ & Times (Seconds)$\downarrow$ \\
    \hline
    3DGS~\cite{kerbl20233dgs}        & -              & -              & -                 & -                           \\
    InstantSplat~\cite{instantsplat} & 32.78          & 0.945          & 0.160             & \textbf{57}                 \\
    CF-3DGS~\cite{cf3dgs}            & 18.37          & 0.803          & 0.32              & 372                         \\
    \textbf{Ours}                    & \textbf{34.50} & \textbf{0.954} & \textbf{0.135}    & 69                          \\
    \hline
  \end{tabular}
  \label{tab:3view}
\end{table}

\begin{figure}[htb]
  \centering
  \includegraphics[width=\textwidth]{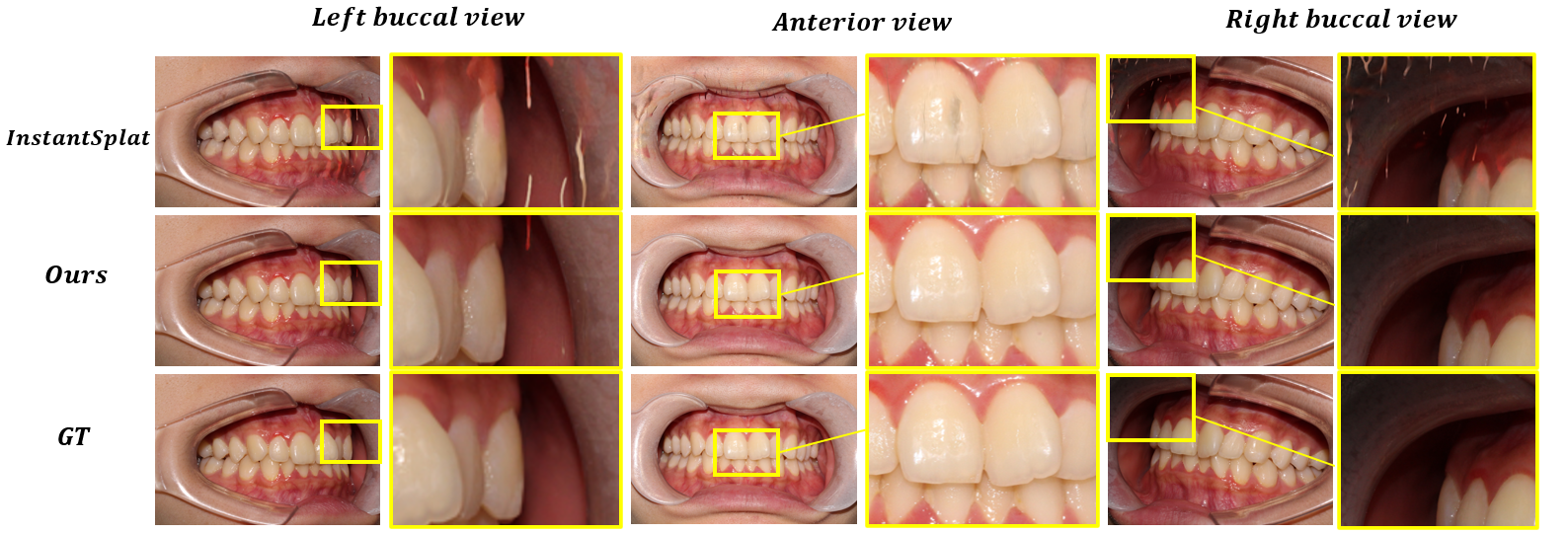}
  \caption{\textbf{Reconstruction Comparisons with 3 views.} Visualization of Rendered Images and GT with 3 views Input.}
  \label{fig:3viewTrain}
\end{figure}

\begin{figure}[htb]
  \centering
  \includegraphics[width=\textwidth]{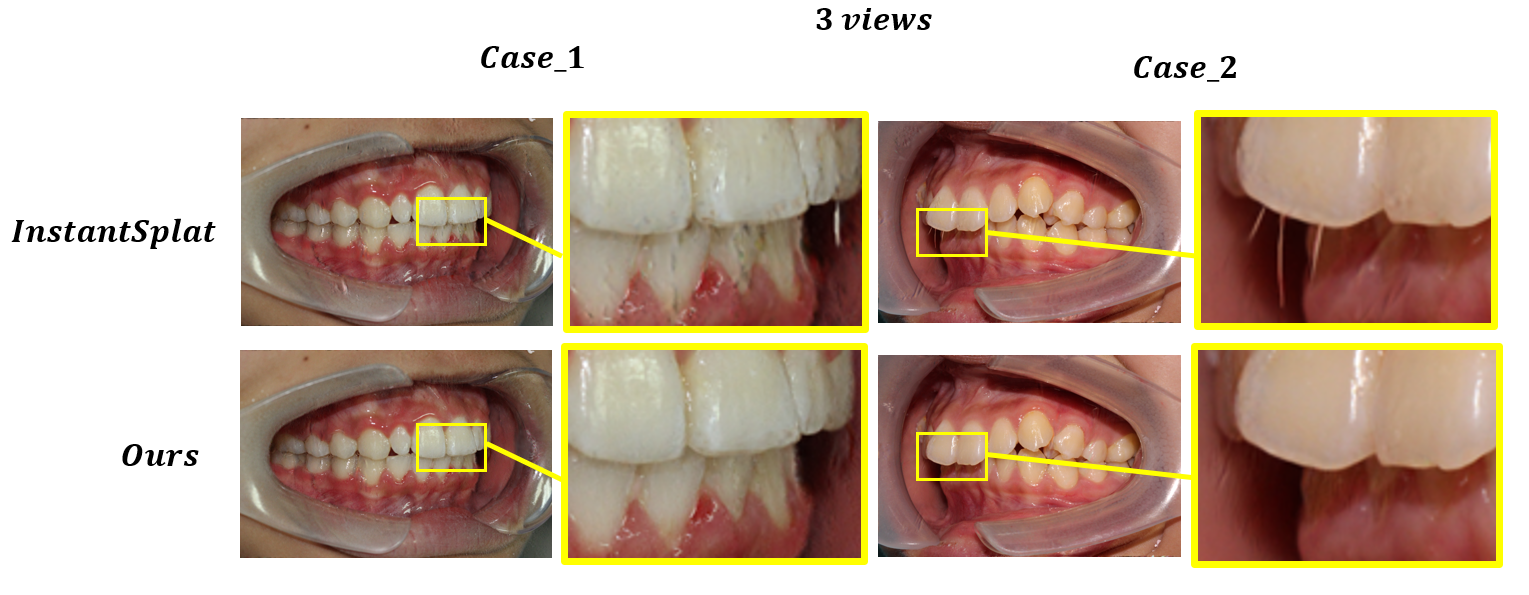}
  \caption{\textbf{Novel View Synthesis Comparisons with 3 views.}
    Due to the lack of ground truth for the 3-view input setting, our analysis focuses on relative performance improvements.}
  \label{fig:3viewNovel}
\end{figure}

For qualitative evaluation, Figure~\ref{fig:3viewTrain} presents the input
training views used in the experiments, and Figure~\ref{fig:3viewNovel}
visualizes synthesized views that were not seen during training. Although
ground truth is unavailable for these novel viewpoints, relative comparisons
indicate that our method successfully reconstructs dental structures with high
fidelity, free from geometric holes or blurring. This confirms the
effectiveness of our framework in producing high-quality reconstructions even
with extremely limited inputs.

\subsection{Ablation Study}
We present the ablation study in Table~\ref{tab:ablation_results} to validate
the contribution of each component in our framework. We conduct an ablation
study on the proposed SAP strategy, optical flow constraint(Flow), gradient
constraint(Gradient), and confidence mask within the flow loss(FlowMask), as
summarized in Table~\ref{tab:ablation_results}.
\begin{table}[htb]
  \centering
  \caption{\textbf{Ablation study} of our method under sparse-view(6 views) setup.}
  \begin{tabular}{c c c c c}
    \hline
    Ablation Setting & PSNR$\uparrow$ & SSIM$\uparrow$ & LPIPS$\downarrow$ & Times (Seconds)$\downarrow$ \\
    \hline
    w.o. Gradient    & 27.58          & 0.857          & 0.288             & 62                          \\
    w.o. FlowMask    & 27.94          & 0.843          & 0.261             & 64                          \\
    w.o. Flow        & 27.35          & 0.852          & 0.285             & \textbf{57}                 \\
    w.o. SAP         & 28.31          & 0.867          & 0.254             & 70                          \\
    Full model       & \textbf{28.41} & \textbf{0.872} & \textbf{0.247}    & 72                          \\
    \hline
  \end{tabular}
  \label{tab:ablation_results}
\end{table}

When the gradient constraint is removed, the performance drops significantly.
This is because the gradient loss plays a key role in guiding the densification
strategy of 3DGS, and its absence hinders effective Gaussian expansion.
Similarly, removing both the optical flow and the associated confidence mask
also leads to a notable decline in performance. However, this configuration
results in a substantial increase in training efficiency, as the geometric
constraints from optical flow introduce additional computational overhead and
increase the number of parameters to optimize. When the SAP strategy is
removed, the performance decreases slightly. This is because the initialization
quality mainly affects the early-stage convergence speed of 3DGS. As training
progresses, the network continuously optimizes the Gaussians through the joint
minimization of photometric, flow, and gradient losses, gradually compensating
for the impact of suboptimal initialization.

\section{Conclusion}
In this paper, we introduce DentalSplat, the first reconstruction framework for
dental occlusion based on Dust3R, capable of supporting dynamic, sparse, and
unposed input images. Extensive experiments with our collected dataset
demonstrate that the incorporated geometric and gradient optimization
strategies are highly effective for orthodontic scenarios, with the quality of
synthesized novel views significantly surpassing that of state-of-the-art
models. For remote orthodontics, the system requires only a video or a few
images to complete scene training and high-quality novel view synthesis within
a minute.

%
%
%
\bibliographystyle{splncs04}
\bibliography{mybibliography}

\end{document}